\documentclass[11pt,a4paper]{article}
\usepackage[hyperref]{acl2020}
\usepackage{graphicx}
\usepackage{times}
\usepackage{latexsym}
\usepackage{amsmath}
\DeclareMathOperator*{\argmax}{argmax}

\usepackage{microtype}
\usepackage{subcaption}
\usepackage{verbatim}
\usepackage{amssymb}
\usepackage{multirow}
\usepackage{arydshln}
\usepackage[warn]{textcomp}
\usepackage{caption}
\usepackage{dirtytalk}
\usepackage[ruled,vlined]{algorithm2e}
\usepackage[toc,page]{appendix}

\aclfinalcopy 

\title{Policy-Driven Neural Response Generation for Knowledge-Grounded Dialogue Systems}

\author{Behnam Hedayatnia \\
  Amazon Alexa AI \\
  \texttt{behnam@amazon.com} \\\And
  Karthik Gopalakrishnan \\
  Amazon Alexa AI \\
  \texttt{karthgop@amazon.com} \\\And
  Seokhwan Kim \\
  Amazon Alexa AI \\
  \texttt{seokhwk@amazon.com} \\
  \\\AND
  Yang Liu \\
  Amazon Alexa AI \\
  \texttt{yangliud@amazon.com} \\\And
  Mihail Eric\\
  Amazon Alexa AI \\
  \texttt{mihaeric@amazon.com} \\\And
  Dilek Hakkani-T\"{u}r\\
  Amazon Alexa AI \\
  \texttt{hakkanit@amazon.com} \\}

\date{}

\begin{document}
\maketitle
\begin{abstract}
Open-domain dialogue systems aim to generate relevant, informative and engaging responses. Seq2seq neural response generation approaches do not have explicit mechanisms to control the content or style of the generated response, and frequently result in uninformative utterances. In this paper, we propose using a dialogue policy to plan the content and style of target responses in the form of an action plan, which includes knowledge sentences related to the dialogue context, targeted dialogue acts, topic information, etc. The attributes within the action plan are obtained by automatically annotating the publicly released Topical-Chat dataset. We condition neural response generators on the action plan which is then realized as target utterances at the turn and sentence levels. We also investigate different dialogue policy models to predict an action plan given the dialogue context. Through automated and human evaluation, we measure the appropriateness of the generated responses and check if the generation models indeed learn to realize the given action plans. We demonstrate that a basic dialogue policy that operates at the sentence level generates better responses in comparison to turn level generation as well as baseline models with no action plan. Additionally the basic dialogue policy has the added effect of controllability.




\end{abstract}

\begin{figure}[t]
\small
\centering
\begin{tabular}{l}
\hline
...\\
{\bf Speaker 1:} Right. Teams do all kinds of things to bother \\
the competition. I've heard of teams having heated \\ 
benches in the winter for themselves but not for the \\
visitors. \\
{\bf Speaker 2:}  I would hate a cold bench. Then again, I \\
wouldn't want to be some place that cold or watching \\
football.\\
\hline
\\[-2mm]
{\bf Speaker 1:}\\
\\[-2mm]
{\em candidate A}: yeah \\ 
\hline
\\[-2mm]
{\em knowledge}: \\
The NFL has no official rule against female players.\\
\\[-2mm]
{\em candidate B}:\\ 
I heard NFL has no official rule against female players.\\
\\[-2mm]
{\em candidate C:} \\
Yeah. I would hate that too.  Do you follow NFL? I heard\\
they have no official rule against female players. \\
\hline
\end{tabular}
\caption{{\em candidate A} is an uninformative response. By grounding on knowledge we get more informative responses i.e.,~{\em candidates B} and~{\em C}.~{\em candidate B} contains only a statement, leading to an abrupt transition.~{\em candidate C} smoothly transitions topics with dialogue acts: feedback, statement, question, and statement.}
\label{fig:example2}
\end{figure}

%

\section{Introduction}
\label{sec:introduction}
Open-domain dialogue systems have typically been modeled using end-to-end approaches, more specifically encoder-decoder 
architectures~\citep{sordoni2015neural, serban2017hierarchical, serban2016building, vinyals2015neural}. These seq2seq models are commonly trained on a maximum likelihood objective, which leads to repetitive and uninformative responses~\cite{wei2017neural}. As seen in Figure~\ref{fig:example2} {\em candidate A} is a typical generic response given the dialogue context. 
In order to deal with this problem, 
previous work proposed grounding generated responses on knowledge sentences related to the dialogue context~\citep{ghazvininejad2018knowledge,yavuz2019deepcopy, zhou2018commonsense, dinan2018wizard, gopalakrishnan2019topical}. 
To improve the diversity of generated responses, others proposed conditioning response generation on 
latent~\citep{serban2016building, serban2017hierarchical, shen2017conditional, zhao2017learning} or discrete
attributes~\citep{sankar2019deep, li2016persona, see2019makes, serban2017hierarchical}. These discrete attributes are typically presented to the decoder at the turn level, and are not associated with a specific segment of the output.

Another issue with seq2seq approaches is that, due to the lack of explicit control mechanisms, the style of these responses does not always match with what would be suggested by user experience experts. For example, the generated response may not acknowledge what the user just said, or may jump to a new topic without first 
introducing it.  Figure~\ref{fig:example2} shows examples of two response candidates with similar content: \textit{candidate C} acknowledges Speaker 2's previous statement and follows up with a question introducing a new topic and statement, in contrast with 
\textit{candidate B} which abruptly transitions into the new topic. 

According to~\cite{schegloff2007sequence} human conversations 
are sequentially organized units. Turns and actions realized within them are related to what came before and affect what 
comes next. Inspired by the previous studies, we propose a policy-driven neural response generation (PD-NRG) approach for open-domain, 
knowledge-grounded dialogue systems. Our motivation for this work is to have a mechanism for open domain conversational systems, i.e., a dialogue policy,
that can enable such higher-level control of generated responses. The dialogue policy provides a sequential organization plan or {\em action plan}. The {\em action plan} specifies the order and relationship of sentences within a turn targeting engaging responses to users throughout the interaction.
This form of control is similar to dialogue management (DM) and natural language generation (NLG) in task-oriented systems 
where a meaning representation determined by the DM is realized as a response during NLG.
To further control the content and order of sentences within the generated response, previous work on task-oriented systems proposed
explicit content and sentence planning~\cite{walker2007individual}. 
Previous work for open-domain dialogue systems also follow a similar method for sentence 
planning and design dialogue policies to predict a set of discrete attributes such as topic and dialogue acts~\citep{fang2018sounding, cervone2017roving, yu2019gunrock, bowden2019slugbot, fulda2018byu, Pichl2018Alquist2A, ahmadvand2018emory}. However, these
studies rely on a set of templates for NLG, resulting in repetitive response structures.

We design a set of dialogue policy models that adapt to the dialogue context to appropriately control the responses at both the turn and sentence-levels. 
We extend the end-to-end-approach of~\cite{dinan2018wizard, gopalakrishnan2019topical}: we take in as input both the dialogue context and an action plan to predict the next response. We train our PD-NRG model by fine-tuning on the Generative Pre-trained Transformer (GPT)~\cite{radford2018improving} model in a TransferTransfo fashion~\cite{wolf2019transfertransfo}. The TransferTransfo model is a state-of-the-art neural open-domain dialogue system that won 1st place in automated evaluation and 2nd place in human evaluation at the NeurIPS ConvAI2 conversational Intelligence Challenge~\cite{dinan2020second}. Our approach differs from previous works that condition on discrete attributes independently by conditioning on these attributes jointly.

Our contributions include: 
\vspace*{-1ex}
\begin{itemize}
\itemsep -0.75ex
\item [i.] an amended version of the Topical-Chat dataset with annotations on multiple attributes (knowledge, topic, dialogue act). These annotations were tagged automatically which reduces the cost and time of manual annotation while still obtaining strong results.\footnote{https://github.com/alexa/Topical-Chat/tree/master/TopicalChatEnriched}
\item[ii.] the design of a basic dialogue policy to predict an action plan for controllable generation for neural response generators
\item[iii.] a sentence-based generation approach that outperforms turn-level generation, and 
\item[iv.] investigation of simple hand-crafted policies as well as automatically learned policies that could be adapted to new applications.
\end{itemize}

\section{Related Work}
\label{sec:related_work}

Controllability of the generated output has been studied for multiple language generation tasks (such as poetry generation and summarization).  Previous work on controlling the style and content of the output of generation focused on two main approaches, conditional generation and weighted decoding. Conditional generation modifies the input to the model to condition the generation on control parameters. For example, for summarization, to control the size of the targeted summary,~\citet{KikuchiEtAl2016} and~\citet{FanEtAl2018} prepended the input to the encoder with the length bin  of the target summary (or its embedding) in sequence-to-sequence models with attention. Similarly, previous works proposed conditioning response generators on latent~\citep{serban2016building, serban2017hierarchical, shen2017conditional, zhao2017learning} or discrete attributes, including dialogue acts~\cite{sankar2019deep}, sentiment~\cite{sankar2019deep}, speaker identifiers~\cite{li2016persona}, lexical features~\cite{see2019makes} or topics~\cite{serban2017hierarchical}. 

Weighted decoding~\cite{see2019makes} instead uses features that are controllable~\citep{GhazvininejadEtAl2017, BahetiEtAl2018} and supplements the scores from the decoder model output with these features. Weighted decoding only allows control with token level attributes, and hence is not ideal for our approach. Instead our work focuses on conditional generation methods with sentence-level control, as described in more detail in Section~\ref{sec:knowledge_grounded_model_action}. Moreover, the goal of these previous studies is to use these attributes to diversify responses for chit-chat generation; whereas our main focus is on knowledge grounded response generation, and how to appropriately include knowledge in a conversation by jointly conditioning our response on a set of control mechanisms. 

There is also previous work on controlling attributes such as question asking at the dialogue level.~\citet{see2019makes} initialized the generation of turns of a dialogue with a fixed distribution that specified what percentage of generated turns should include questions during the dialogue. However this does not allow for flexible control where the number of questions may need to vary depending on the course of the dialogue. A developer may not know the distribution of dialogue acts to use; therefore we focus on learning a dialogue policy model that automatically learns the style of the response based on the dialogue context.

Similar to previous work for response generation we ground our generated responses on knowledge. ~\citet{ghazvininejad2018knowledge} used end-to-end memory networks to represent knowledge,~\citet{yavuz2019deepcopy} added a copy mechanism to attend over the knowledge, and~\citet{zhou2018commonsense} used a static graph attention mechanism for relevant knowledge graphs. ~\citet{dinan2018wizard} and ~\citet{gopalakrishnan2019topical} used memory networks based on transformer architectures~\cite{transformer} to encode knowledge sentences and dialogue history to decode a response.~\citep{roller2020recipes,smith2020can} extended the incorporation of knowledge by creating a dataset, Blended Skills Dataset (BSD), which incorporated multiple skills (personality, knowledge and empathy) across the conversation by combining utterances from existing datasets (PersonaChat~\cite{zhang2018personalizing}, Wizard of Wikipedia~\cite{dinan2018wizard}, EmpatheticDialogues~\cite{rashkin2018towards}). A turn within BSD can convey a value such as sadness from empathy or talk about a specific topic from its knowledge.~\citet{roller2020recipes} trained a memory network based transformer architecture on this dataset. However other than knowledge their model does not explicitly predict the values for all skills at every turn. In this work we integrate knowledge in responses by jointly conditioning on the attributes in the input action plan. The values for these attributes are predicted at every turn based on a dialogue policy.

Previous work also investigated content and sentence planning in open-domain dialogue systems.~\citet{fang2018sounding} used a hierarchical dialogue policy where the top-level decides the topic to talk about followed by prediction of a set of speech acts to generate and combine phrases.~\citet{ahmadvand2018emory} extracts multiple features such as topic, intent, entities, sentiment to send to a dialogue manager to select a response.~\citet{Pichl2018Alquist2A} used a Hybrid Code Network for their dialogue manager and used sentiment, dialogue act and topic to predict which response to select.~\citet{cervone2017roving} segmented an utterance into functional units where each unit is attributed to one of the ISO-standard dialogue acts~\cite{mezza2018iso}.~\citet{yu2019gunrock} learned a segmentation model to break the utterances into smaller segments and predict the topic and dialogue act of each segment.~\citet{bowden2019slugbot} proposed modeling discourse coherence between generated turns.~\citet{fulda2018byu} planned and composed a set of responses such as informative and follow-up questions. However these previous works generated responses from a set of templates which are usually repetitive for open-domain conversations. Our work focuses on neural generative models for response generation in open-domain dialogue systems.

The closest work to ours in terms of learning a dialogue policy for open-domain dialogue is~\cite{xu2018towards} who designed a policy network to predict dialogue acts and fed those acts into a response generation model to control responses. However a key part of open-domain dialogue is to introduce knowledge into a conversation. We design a policy that integrates knowledge with dialogue acts at a sentence-level. In contrast to~\cite{xu2018towards} who used a machine learning based approach, we show that a basic rule-based dialogue policy can result in strong performance.

\section{Dialogue Policy}
\begin{figure*}[t]
    \centering
    \includegraphics[scale=0.5]{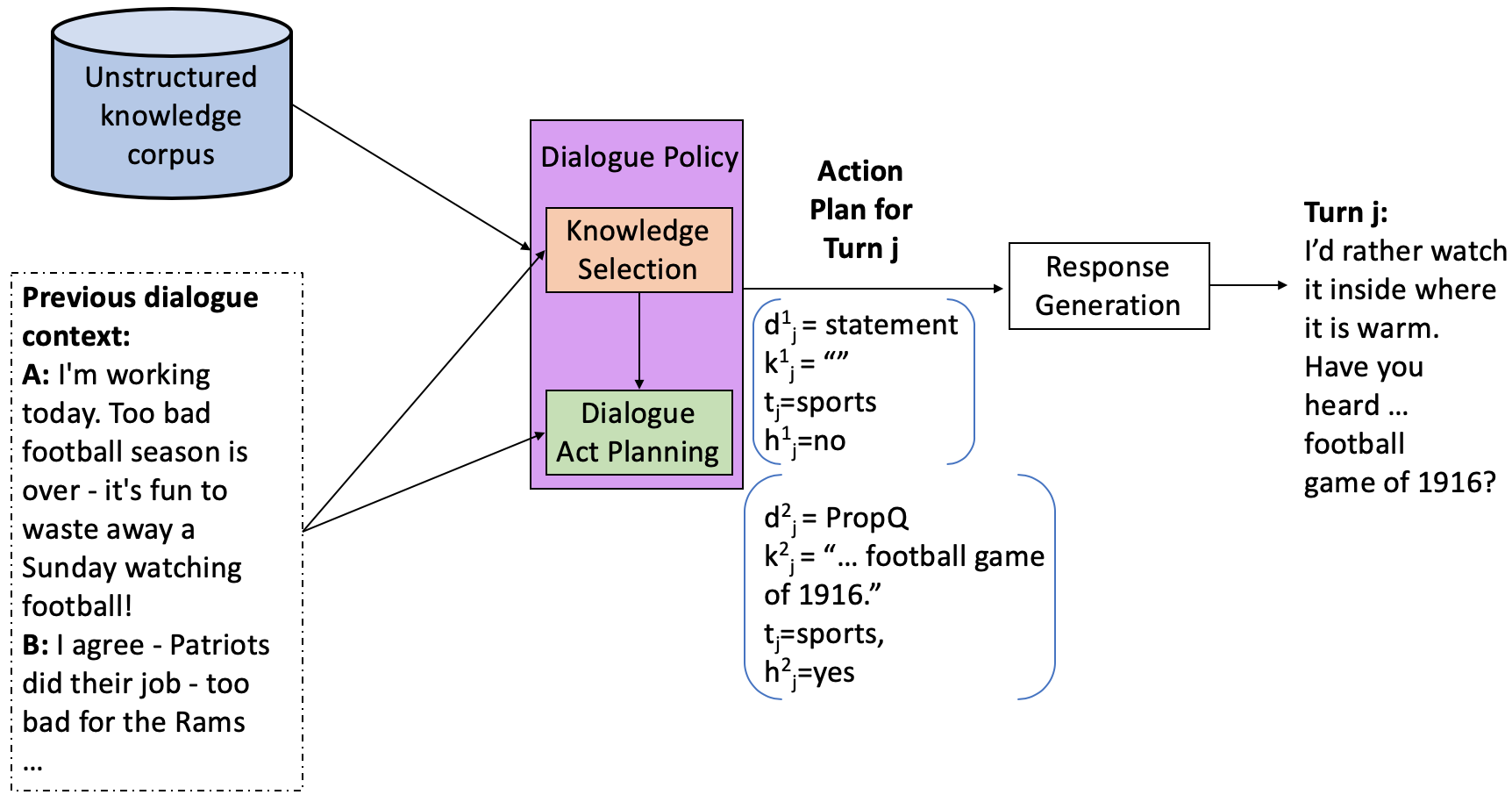}
    \caption{Policy-driven neural response generation.}
    \label{fig:PD-NRG}
\end{figure*}

\vspace{-2.3mm}
\label{sec:estimate_action_plan}
Our proposed PD-NRG approach has two parts: a~\textit{dialogue policy} that determines the action plan based on the dialogue context, and a ~\textit{response generation} model that takes the action plan and the dialogue context as input to generate a response. The dialogue policy has components that predict the individual elements of the action plan:~\textit{knowledge selection} and~\textit{dialogue act planning}. Knowledge selection determines the knowledge to be integrated in the response by finding sentences from a knowledge document corpus that are relevant to the dialogue context. Dialogue act (DA) planning determines the style of the response in the form of dialogue acts to be realized. We have two forms of dialogue act planning methods:~\textit{Knowledge-dependent DA planning} and~\textit{Knowledge-independent DA planning}. Figure~\ref{fig:PD-NRG} depicts the architecture of PD-NRG.

\begin{table}
\vspace*{-1ex}
\centering
\small
\begin{tabular}{lc}
\hline
{\bf Dialogue Act} & {\bf Definition}  \\
\hline
Apology & apology \\
ChoiceQ & Or-question \\
Commissive & Offer, Commit\\ 
Directive & Open-Option, Suggest\\ 
Feedback & Acknowledge, Feedback \\ 
PropQ & Yes-no-question, Suggest\\ 
Salutation & bye, greet\\ 
SetQ & Wh-question\\ 
Statement & Inform \\ 
Thanking & thanking, your-welcome\\ 
\end{tabular}
 \vspace*{-2ex}
\caption{\label{da_table} The subset of ISO-Standard dialogue acts proposed by~\cite{mezza2018iso}.
}
 \vspace*{-2ex}
\end{table}

\subsection{Action Plan (AP)}
\label{sec:action_plan}
For the rest of this work, let $D_j = [x_1, \dots, x_j]$ denote a partial dialogue containing a sequence of $j$ turns. And let $x_i$ represent a turn in a dialogue where $1 \leq i \leq j$. Each $x_i$ contains a sequence of $n_i$ sentences, $x_i = [s_i^1, \dots, s_i^{n_i}]$.

Each $x_i$ also contains an action plan that consists of one frame for each sentence $[f_i^1, \dots, f_i^{n_i}]$. The frames, formed of attributes and values, may include:
\vspace*{-1ex}
\begin{enumerate}
\itemsep -1ex
\item Dialogue acts ($d$) at a sentence-level to help control the style of the generated response. Table~\ref{da_table} lists all the dialogue acts used in this work.
\item Topics ($t$) at a turn-level to generate topically coherent responses. The complete list of topics are: fashion, politics, books, sports, general-entertainment, music, science \& technology and movies.
\item Knowledge ($k$) at a turn or sentence-level to generate interesting and informative responses. The knowledge is represented as a sentence drawn from an unstructured knowledge corpus.
\item Use-knowledge flag ($h$) that signals whether or not to use the knowledge attribute ($k$) at the turn or sentence-level.
\end{enumerate}

Each frame in the action plan corresponds to a sentence and the frame corresponding to a sentence, $s_i$, is denoted as a tuple containing a set of the 4 attributes, ($d_i, t_i, k_i, h_i$). In this work, we focus on these attributes for action plans, as they are the most basic and critical ways to control knowledge-grounded response generation. 

\subsection{Knowledge Selection}
\label{knowledge_selection_gt}

For the knowledge selection component of our dialogue policy, referenced in Figure~\ref{fig:PD-NRG}, we compute the following for each turn $x_i$ at run time. Let $c_i$ be defined as the dialogue history $x_1,..., x_{i-1}$:
\begin{equation}
    \hat{k} = \argmax_{k_m \in K} \left( \frac{\vec{c_{i}} \cdot \vec{k_m}}{\left\lVert \vec{c_{i}} \right\rVert \left\lVert \vec{k_m} \right\rVert} \right)
    \label{cosine_similarity}
\end{equation}
$k_m$ is a knowledge sentence from an unstructured knowledge corpus, $K$,  in the Topical-Chat dataset~\cite{gopalakrishnan2019topical}. We use the BM25 model~\cite{robertson2009probabilistic} to rank knowledge sentences and represent
$\vec{c_{i}}$ and $\vec{k_m}$ as TF-IDF vectors for $c_{i}$ and $k_m$. We compute cosine similarity between the vectors and~\textit{argmax} over the all $k_m$ in our knowledge corpus. For $c_{i}$ we are only using the most recent previous turn $x_{i-1}$ for selection.
To decide whether or not to use knowledge, we manually set a threshold value of 0.2 on the similarity score between the sentences. If the similarity score is above the threshold we use the knowledge sentence as input and vice versa.

\subsection{Dialogue Act planning}
For dialogue act planning, we define a set of dialogue act transitions from common examples in the Topical-Chat corpus.
The set of dialogue acts for the next response are determined by both dialogue acts and the knowledge sentence selected, based on the dialogue context. Figure~\ref{fig:PD-NRG} shows the output of the knowledge selection being fed as input into the dialogue act planning component. We represent the transitions as a decision tree. Based on which set of dialogue acts were outputted, we decide whether or not to include the knowledge sentence. Some dialogue acts, such as {\em Feedback}, do not need to include knowledge by definition.

\subsubsection{Knowledge-dependent DA planning}
\label{sec:complex_policy}
We propose a Knowledge-dependent DA planning (KD-DA-P) where there are two inputs to predict the dialogue acts for the current turn $x_{j+1}$:
\vspace*{-1ex}
\begin{enumerate}
    \itemsep -1ex
    \item the last dialogue act associated with the previous sentence $s_j^{n_j}$
    \item the output of knowledge selection
\end{enumerate}

The dialogue act model looks at the output of the knowledge selection model to see if the knowledge selected is the same or different as compared to the knowledge sentence selected for the previous turn $x_j$. Based on this information a certain subset of the transitions defined for dialogue act planning are used to predict the dialogue acts for the next response. We represent the KD-DA-P as a decision tree, and include it in the appendix.

\subsubsection{Knowledge-independent DA planning}
The prediction of the dialogue acts is done independently of the selected knowledge in four ways:
\begin{enumerate}
    \itemsep -0.5ex
    \item {\em Simple DA planning:} We define a set of transitions that determine the set of dialogue acts for the next response based solely on the previous dialogue acts. We represent the transitions as a decision tree, and present it in the appendix.
    \item {\em Seq2Seq Model for DA planning:} Using the OpenNMT library~\cite{opennmt}, we train a sequence-to-sequence model based on bi-directional LSTMs with Luong attention~\cite{luong2015effective} to estimate the dialogue acts of the current turn given the dialogue context $D_j$. During training, each dialog act label is a separate token in the vocabulary and has its own embedding vector. Both the dialogue act and word embeddings are initialized randomly and learned during training.
    \item {\em PropQ DA planning:} For comparison to previous work we use the method in~\cite{see2019makes} which initializes the distribution of questions to be asked at the beginning of the conversation. The work finds that the best model generates questions 65.7\% of the time. At each time-step the PropQ dialogue act is picked 65.7\% of the time thereby replicating this baseline. As shown in Table~\ref{da_table} PropQ corresponds to a Yes-No question which is the most represented question dialogue act in our dataset. We represent the transitions as a decision tree, and present it in the appendix.
    \item {\em AllQ DA planning:} We extend the PropQ DA Prediction baseline above by selecting the PropQ, ChoiceQ or SetQ questions each 21.9\% of the time summing up to 65.7\%.~\citet{see2019makes} does not make a distinction as to what type of questions were asked. We represent the transitions as a decision tree, and present it in the appendix.
    
\end{enumerate}

\section{Policy-driven Response Generation}
\label{sec:knowledge_grounded_model_action}
\vspace{-2.2mm}
As shown in Figure~\ref{fig:PD-NRG}, at a given turn in the dialogue context, the goal of the response generator is to realize the action plan output by the dialogue policy. Our proposed models generate the next turn based on the action plan at a sentence-level, in a sequential manner as opposed to at a turn-level. As shown in Figure~\ref{fig:input_fed} when decoding/generating each sentence of the next turn, the dialogue context $D_j$ as well as the previous sentences generated for the next turn till that iteration are used as input. Algorithm~\ref{sentence_generation} shows the process for sentence-level generation. As seen in the algorithm all the attributes within the AP are jointly taken in as input. 
To jointly condition on the action plan, each attribute is concatenated to the dialogue history as shown in Figure~\ref{fig:input}. 
In the training process each dialog act label is a separate token in the vocabulary and has its own embedding vector which is initialized randomly and learned during training. 
To train our model we represent the knowledge sentence and topic label with the pretrained embeddings from the GPT model whose vocabulary is BPE tokenized. Finally, the use-knowledge flag decides whether or not to include the knowledge embeddings as part of the input. In some of our experiments, we also include the dialogue acts for the past turns by concatenating each turn in the dialogue history with its respective acts.


\newcommand\mycommfont[1]{\footnotesize\ttfamily\textcolor{blue}{#1}}
\SetCommentSty{mycommfont}

\begin{algorithm}
\SetAlgoLined
\KwResult{$x_{j+1}$}
 Given $D_j$ \\
 $x_{j+1}$ = [] \\
 ActionPlan = $[f_{j+1}^1, \dots, f_{j+1}^{n_{j+1}}]$ \\
 for idx in range(len(ActionPlan)) \\
 $~~~$ $f$ = ActionPlan[idx] \\
 $~~~$ $y$ = Model($D_j$,~$f$) \\
 $~~~$ $x_{j+1}$ = $x_{j+1}$ $\oplus$~$y$ \\ 
 $~~~$ \tcp{the step below ensures the}
 $~~~$ \tcp{newly generated sentence is}
 $~~~$ \tcp{included as part of the}
 $~~~$ \tcp{dialogue context for the}
 $~~~$ \tcp{generation of the following}
 $~~~$ \tcp{sentence}
 $~~~$ $D_j$ = $D_j$ $\oplus$~$y$ \\
 return $x_{j+1}$ \\
 \caption{Sentence-level generation}
 \label{sentence_generation}
\end{algorithm}





\subsection{Models for response generation}
\begin{figure*}
\begin{subfigure}[t]{0.4\textwidth}
\hspace*{-0.5cm} 
\includegraphics[scale=0.5]{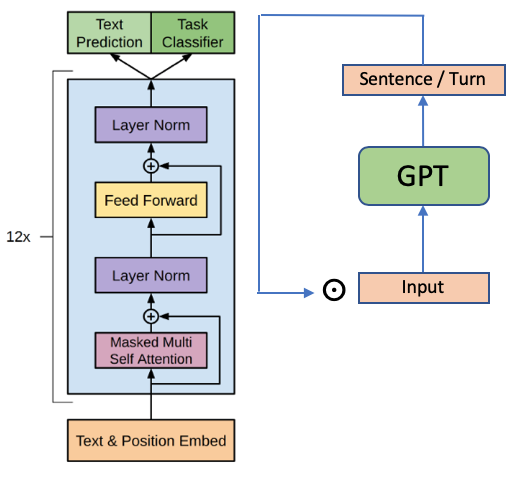}
\caption{GPT Model~\cite{radford2018improving}}
\label{fig:input_fed}
\end{subfigure}\qquad
\begin{subfigure}[b]{0.4\textwidth}
\hspace*{-0.5cm}
\includegraphics[scale=0.62]{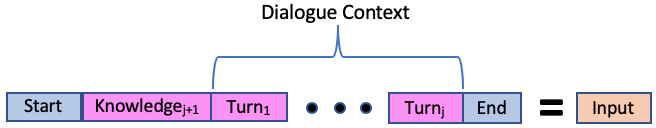}
\caption{Input into Baseline-Turn Model}
\label{fig:input_turn}
\vspace*{0.5cm}

\hspace*{-3.2cm} 
\includegraphics[scale=0.62]{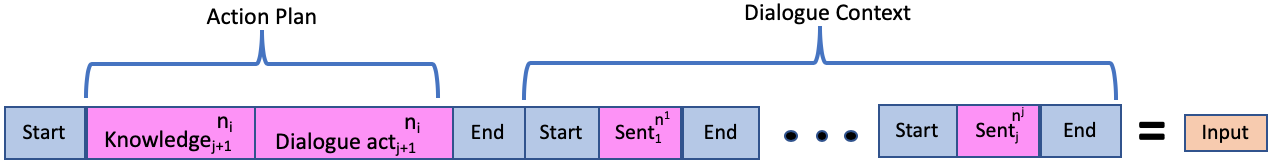}
\caption{Input into PD-NRG Model}
\label{fig:input}
\vspace*{-0.5cm}
\end{subfigure}
\caption{Input for Baseline-Turn Model and PD-NRG model respectively. Figure~\ref{fig:input_fed} shows the generation process where the input is fed into the GPT model. The output is then concatenated back to the input. This process repeats until generation is complete.}
\label{fig:input_models}
\end{figure*}

For all our models, we use the Generative Pre-trained Transformer (GPT)~\cite{radford2018improving} model to finetune in a TransferTransfo fashion~\cite{wolf2019transfertransfo}. The TransferTransfo model is a state-of-the-art neural open-domain dialogue system that won 1st place in automated evaluation and 2nd place in human evaluation at the NeurIPS ConvAI2 conversational Intelligence Challenge~\cite{dinan2020second}. 
\begin{itemize}
\itemsep -0.5ex
\item {\bf Model for turn-level generation:}
\label{sec:baseline_1}
As depicted in Figure~\ref{fig:input_turn}, our baseline~\cite{wolf2019transfertransfo} is given the dialogue context and knowledge sentence as input and predicts the response at the turn-level. 

\item {\bf Models for sentence-level generation:}
As depicted in Figure~\ref{fig:input}, the PD-NRG models are given the action plan (AP) and the dialogue context as input to perform sentence-level prediction. Table~\ref{model_table} lists the versions of PD-NRG models we experimented with along with their corresponding APs. Baseline-Sent is similar to the Baseline-Turn model, except it generates responses sentence-by-sentence. The model generates as many sentences as in the human response. 
\end{itemize}

\begin{table}[ht]
\vspace*{-1ex}
\centering
\begin{tabular}{lclclc|c}
\hline
PD-NRG Models  & Action Plan (AP)\\
\hline
w/ DA & $\{d_{j+1}^{m}, k_{j+1}\}$ \\ 
+knowledge flag & $\{d_{j+1}^{m}, k_{j+1}^{m}, h\}$  \\ 
+knowledge flag +topic & $\{d_{j+1}^{m}, k_{j+1}^{m}, t_{j+1}, h\}$ \\ 
\hline
Baseline Models\\
\hline
Baseline-Turn & $\{k_{j+1}$\} \\
Baseline-Sent & \{$k_{j+1}$\}\\
\end{tabular}
 \vspace*{-2ex}
\caption{\label{model_table}Models and their input AP for every timestep $m$ where $1 \leq m \leq n_{j+1}$.}
 \vspace*{-2ex}
\end{table}

\section{Experiments and Evaluation}
\label{sec:experimental_setup}

\subsection{Dataset}
\label{sec:dataset}
We use the publicly released Topical-Chat\footnote{https://github.com/alexa/Topical-Chat} dataset, a large and diverse knowledge-grounded open-domain dialogue dataset where the underlying knowledge spans 8 broad topics including fashion, books, and so on~\cite{gopalakrishnan2019topical}. Each dialogue contains 20+ turns alternating between two crowd workers. For each dialogue there is a reading set for each crowd worker. Each reading set has three entities and a set of corresponding knowledge sentences. When presenting the results, we use both test sets provided with the corpus, test frequent and test rare. Frequent and rare refer to the frequency of topics and entities being discussed in the training set.

\vspace*{-2ex}
\subsection{Annotating attributes in Topical-Chat}
\label{sec:annotation}
The dataset does not have annotations for some attributes such as dialogue acts or fine-grained associations between knowledge sentences and dialogue turns. Hence, we used out-of-the-box or simple models to automatically annotate our dataset with each attribute, as defined in Section~\ref{sec:action_plan}. The out-of-the-box model use is the SVM tagger released by ~\cite{mezza2018iso}, which can be found at https://github.com/ColingPaper2018/DialogueAct-Tagger. We assume these annotations are the ground-truth attributes for the ground-truth action plan and use them for testing controllability without degrading the response appropriateness. By automatically annotating we reduce the cost and time it takes to manually annotate our dataset along with getting strong results. We measure both the controllability and appropriateness with human evaluation. 

\vspace*{-1ex}
\subsubsection{Annotating Knowledge Sentences}
\label{sec:oracle_knowledge}
Each conversation in Topical-Chat has a pair of reading sets that were presented to crowdworkers before the conversation, to have a knowledgeable interaction. During their conversation crowd workers are asked to annotate which topics/entities were attributed to their turns in the conversation. However there is no fine-grained annotation of which knowledge sentence or sentences were used for a turn, hence we create ground-truth knowledge annotations as a corpus post-processing step. To obtain the knowledge annotation for each turn, we compute similarity between $x_{j+1}$ and $k_m$ using Equation~\ref{cosine_similarity}. We obtain the knowledge annotation for each sentence within a turn by computing similarity between $s_{j+1}^{n_i}$ and $k_m$ using Equation~\ref{cosine_similarity}. To obtain the knowledge annotation for each sentence within a turn, we tokenize the turn into individual sentences. For sentence-tokenization we use the NLTK library~\cite{loper2002nltk}. 

We decide whether or not the turn or sentences within a turn should be
linked to a knowledge sentence by manually setting a threshold value on the similarity score between the knowledge and turn or sentences within a turn. We use the same threshold, 0.2, as described in Section~\ref{knowledge_selection_gt}.


\begin{figure}[ht]
 	\centering
 		\includegraphics[scale=0.5]{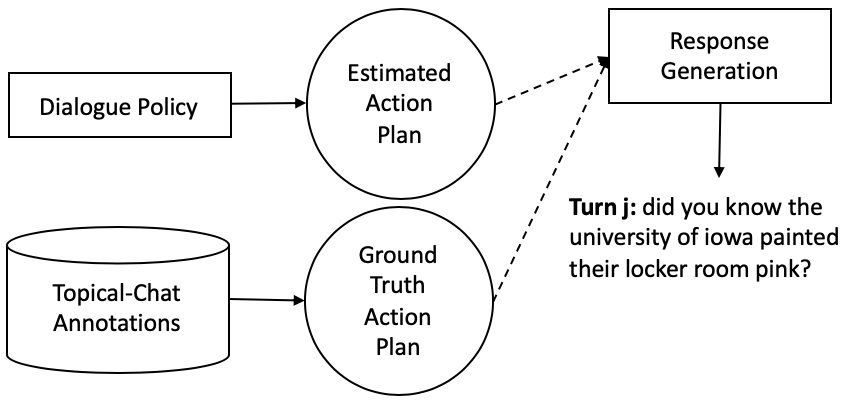}
 	\caption{We calculate automated metrics with both a ground truth and an estimated action plan}
	\label{fig:gt_estimate}
\end{figure}

\subsubsection{Annotating Dialogue Acts}
\label{sec:oracle_dialogue_act}
We obtain the dialogue acts for each sentence by running an off-the-shelf SVM dialogue act tagger\footnote{https://github.com/ColingPaper2018/DialogueAct-Tagger}~\cite{mezza2018iso} which takes in as input the current sentence to predict one of 11 dialogue acts listed in Table~\ref{da_table}. We also experimented with using past dialogue acts predicted from the tagger as additional input; however, this did not change the result. If the confidence score from the SVM tagger is not above a threshold of 0.5, the tagger would output no dialogue act which we denote with a special dialogue act token {\em NoDialogueAct}. 2.1\% of sentences within the Topical-Chat dataset were labeled as~\textit{NoDialogAct}. We assume these are the ground-truth dialogue acts in our dataset. To view the performance of the model we ask two crowd workers to segment and annotate a small set of 100 turns into individual sentences along with their respective dialogue act. The dialogue act tagger obtained an F1 of 0.54, precision of 0.77 and a recall of 0.59 on consolidated test set. 

\vspace*{-1ex}
\subsubsection{Annotating Topic Labels}
\label{sec:oracle_topic}
For the topic label, we use the topic annotations by the Turkers from the original Topical-Chat data collection. For each turn there are multiple topic annotations; however, unlike the dialogue acts and knowledge sentence, topic annotations are at the turn level and are not linked to individual sentences.



\begin{table*}[ht]
  \centering
  \small
\begin{tabular}{l c c c c c c c c}
& & & & & \multicolumn{2}{c}{Avg \#}\\
Models & Past DA & PPL & BLEU-4@1 & ROUGE-L & words & sentences\\
\hline
Human &  & - / - & - / - & - / - & 24.3 / 25.0 & 2.10 / 2.15 \\
Baseline-Turn~\cite{wolf2019transfertransfo} & & 12.92 / 13.53  & 0.024 / 0.028 & 0.134 / 0.130 & 20.7 / 21.7 & 1.87 / 1.93 \\
\hline
Baseline-Sent & & 13.85 / 14.36  & 0.016 / 0.021 & 0.107 / 0.103 & 13.7 / 13.9 & 2.09 / 2.15 \\
PD-NRG w/ DA & & 12.72 / 13.01 & 0.024 / 0.027 & 0.121 / 0.118 & 18.5 / 19.3 & 2.05 / 2.10 \\
PD-NRG w/ DA & \checkmark & 12.39 / 12.80 & 0.021 / 0.021 & 0.115 / 0.111 & 16.0 / 15.8 & 1.77 / 1.77  \\
+knowledge flag & & 12.66 / 12.99 & 0.025 / 0.027 & 0.122 / 0.118 & 17.3 / 18.1 & 2.03 / 2.08 \\
+knowledge flag & \checkmark & \textbf{12.25} / \textbf{12.62} & 0.019 / 0.020  & 0.113 / 0.108  & 15.2 / 15.3 & 1.68 / 1.76\\
+knowledge flag +topic & & 12.76  / 13.07 & 0.023 / 0.026 & 0.123 / 0.117 & 16.8 / 18.2 & 2.10 / 2.14  \\
+knowledge flag +topic & \checkmark & 12.28 / 12.65 & 0.019 / 0.020 & 0.115 / 0.109 & 16.3 / 16.7 & 1.82 / 1.85\\

& & & & & \multicolumn{2}{c}{Corpus Diversity} \\
& & F1 & Precision & Recall & ~\textit{n}=1 & ~\textit{n}=2 \\
\hline
Human & & - / -  & - / - & - / - & 0.037 / 0.050 & 0.266 / 0.326 \\
Baseline-Turn~\cite{wolf2019transfertransfo} & & \textbf{0.249} / \textbf{0.253} & 0.275 / 0.272 & 0.229 / 0.236 & 0.018 / 0.027 & 0.118 / 0.165 \\
\hline
Baseline-Sent & & 0.220 / 0.220 & 0.258 / 0.252 & 0.191 / 0.195 & 0.018 / 0.026 &  0.115 / 0.156 \\
PD-NRG w/ DA & & 0.241 / 0.240 & 0.281 / 0.279 & 0.210 / 0.212 & 0.018 / 0.027 & 0.123 / 0.165 \\ 
PD-NRG w/ DA & \checkmark & 0.230 / 0.227 & 0.291 / 0.287 & 0.185 / 0.185 & 0.021 / \textbf{0.032} & 0.133 / 0.180 \\
+knowledge flag & & 0.240 / 0.242 & 0.280 / 0.280 & 0.209 /  0.213 & 0.018 / 0.027 & 0.122 / 0.164 \\
+knowledge flag & \checkmark & 0.222 / 0.223 & 0.287 / 0.281 & 0.180 / 0.180 & \textbf{0.032} / \textbf{0.022} & \textbf{0.137} / \textbf{0.181} \\
+knowledge flag + topic & & 0.244 / 0.245 & 0.276 / 0.274 & 0.210 / 0.213 & 0.018 / 0.027 & 0.118 / 0.159  \\
+knowledge flag + topic & \checkmark & 0.224 / 0.221 & 0.272 / 0.271 & 0.187 / 0.186 & 0.020 / 0.029 & 0.136 / 0.177  \\
\end{tabular}
 \vspace*{-2ex}
\caption{\label{auto_relevancy_diversity}
Automated metrics with ground-truth Action Plan on test freq / rare
 \vspace*{-2ex}
}
\end{table*}

\vspace*{-2ex}
\subsection{Evaluation Measures}
For automatic evaluation we compute a set of metrics between our generated and ground truth response: perplexity, BLEU-1, ROUGE-L, unigram F1-score. We also compute n-gram diversity as defined in~\cite{ghazvininejad2018knowledge}.

For human evaluation, we followed a similar setup as~\cite{li2016deep} and generated 200 snippets which contain a dialogue context of 5 turns. We generated responses from 2 models to compare against. We asked a set of 3 crowd workers ``Which final response is more appropriate for the given conversation?". Our MTurk layout for evaluation is shown in the Appendix.

\subsection{Results using the Ground-Truth Action Plan}
We first check whether the PD-NRG approach results in better responses when we use the ground truth action plan. As seen in Figure~\ref{fig:gt_estimate} instead of using a dialogue policy, we form ground truth action plans from the annotations described in Section~\ref{sec:annotation}. We then use them to generate a response for that turn. 
Table~\ref{auto_relevancy_diversity} presents automated evaluation results for Baseline-Turn, Baseline-Sent and variations of the PD-NRG models.
As seen in the results table, adding dialogue acts increases diversity for all the proposed models. This aligns with previous work that using dialogue acts leads to more diverse responses~\cite{sankar2019deep}. The F1-scores of the PD-NRG w/ DA model are lower than the Baseline-Turn model due to the PD-NRG model decoding shorter sentences resulting in lower recall.
The PD-NRG w/ DA model with the addition of previous dialogue acts as input results in the lowest perplexity for both frequent and rare test sets.

\subsubsection{Do the Models follow the Action Plan?}
\label{sec:human_results_realize}
By jointly conditioning on the attributes in the action plan, we aim to control multiple aspects of the response, such as content and style. The dialogue acts determine if the response should be a question, statement or should give feedback. The knowledge determines what content should be present in the response. 
To see if the model responses follow the action plan, we manually annotated if the model's responses realize the dialogue acts and their respective knowledge sentence (focusing on the cases where the action plan included a knowledge sentence) in their input. Turns with no dialogue acts, i.e.,  marked as~\textit{NoDialogAct}, were ignored. The results from the manual evaluation are presented in Table~\ref{human_evaluation_rank}. The PD-NRG w/ DA + knowledge flag model has the highest accuracy in realizing the input action plan, achieving 80.6\% accuracy on the dialogue acts of the generated responses, and 52.1\% accuracy in correctly integrating the provided knowledge sentences. Figure~\ref{fig:example} presents an example from this model.

\begin{table}
\centering
\small
\begin{tabular}{l c c@{\hspace{0.1cm}} c}
Models & Past DA & \% DA & \%K\\
\hline
Baseline-Turn~\cite{wolf2019transfertransfo} & & 26.7 & - \\
\hline
Baseline-Sent & & 59.4 & 30.8 \\
PD-NRG w/ DA & & 69.1 & 47 \\
PD-NRG w/ DA & \checkmark & 69.7 & 39.3 \\
+knowledge flag & & \textbf{80.6} & \textbf{52.1}\\
+knowledge flag & \checkmark &68.1 & 47.8 \\
+knowledge flag +topic & & 77.8 & 47.4\\
+knowledge flag +topic & \checkmark & 69.0 & 45.3\\
\hline
\end{tabular}
\vspace*{-2ex}
\caption{\% of Dialogue Acts (DA) and Knowledge (K) Realized for PD-NRG Models to showcase controllability.
\label{human_evaluation_rank}}
 \vspace*{-2ex}
\end{table}

\begin{figure}[h]
\small
\centering
\begin{tabular}{l}
...\\
{\bf Speaker 1:} Free with you, they should have had Snoop Dogg \\ 
make a theme song for the game like  \\
he did for his son's high school football team LOL \\
{\bf Speaker 2:} Interesting, do you play golf? \\
\hline
\\[-2mm]
{\bf Speaker 1:}\\
\\[-2mm]
{\em Baseline-Turn}:\\ 
no, i don't play golf, but i hear \\ 
it has been a lot of years since the last time.\\
\\[-2mm]
{\em PD-NRG model:} \\
Statement \textrightarrow{} not really, i'm not a huge fan of golf. \\
PropQ  \textrightarrow{} have you ever played? \\
\hline
\end{tabular}
\caption{Baseline-Turn model versus PD-NRG model}
\label{fig:example}
\end{figure}

\begin{table}[h]
\centering
\small
\begin{tabular}{l@{\hspace{0.1cm}} c c@{\hspace{0.2cm}} c }
& & \multicolumn{2}{c}{Avg \#}\\
Policy & F1 & words & sentences\\
\cline{0-3}
Ground truth & 0.22 / 0.22 & 15.2 / 15.3 & 1.68 / 1.76\\
Baseline-Turn & 0.18 / 0.17  & 19.8 / 19.7 & 1.86 / 1.87\\
\cline{0-3}

KI-DA-P (Simple) & 0.14 / 0.14 & 12.9 / 12.2 & 1.89 / 1.89 \\

KD-DA-P & 0.14 / 0.14 & 12.3 / 11.5 & 1.91 / 1.91 \\

KI-DA-P(Seq2Seq) & 0.14 / 0.17 & 13.1 / 13.4 & 1.46 / 1.56 \\

\end{tabular}
\vspace*{-2ex}
\caption{\label{auto_predicted_relevant_diversity}
Automated metrics with estimated Action Plan. Baseline-Turn~\cite{wolf2019transfertransfo}}
\vspace*{-2ex}
\end{table}

\begin{table}
\centering
\small
\begin{tabular}{l@{\hspace{0.1cm}} c@{\hspace{0.2cm}}c@{\hspace{0.2cm}}c@{\hspace{0.2cm}}c}
 \hline
Dialogue policy & \%W & \%T &\%L & IAA \\
\hline
KD-DA-P vs. B-Turn*  & 40.8 & 30.3 & 28.9 & 0.43  \\
KI-DA-P(Seq2Seq) vs. B-Turn* & 25.1 & 35.7 & 39.2 & 0.47  \\
KD-DA-P vs. KI-DA-P(PropQ)** & 54.2 & 5.5 & 40.2 & 0.46 \\
KD-DA-P vs. KI-DA-P(AllQ)** & 54.1 & 7.4 & 38.3 & 0.48\\
KD-DA-P vs. Human response**  & 16.7 &35.3 & 48.0 & 0.53  \\
\hline
\end{tabular}
\vspace*{-2ex}
\caption{\% of Wins(W), Ties (T) and Losses(L) for the baseline models vs PD-NRG model on appropriateness. The KD-DA-P policy is statistically significant compared to the B-Turn(Baseline-Turn)~\cite{wolf2019transfertransfo} as well as the KI-DA-P(PropQ) and KI-DA-P(PropQ) baselines~\cite{see2019makes}. We compute Krippendorff's alpha for Inter-annotator agreement(IAA). We computed the p-value using a two-tailed binomial test. * refers to a p-value $<$ 0.05 and ** refers to a p-value $<$ 0.01. 
\label{human_evaluation_policy_turn}}
 \vspace*{-2ex}
\end{table}


\subsection{Results using an estimated Action Plan}
\begin{figure}[t]
  		\centering
  		\includegraphics[width=\columnwidth]{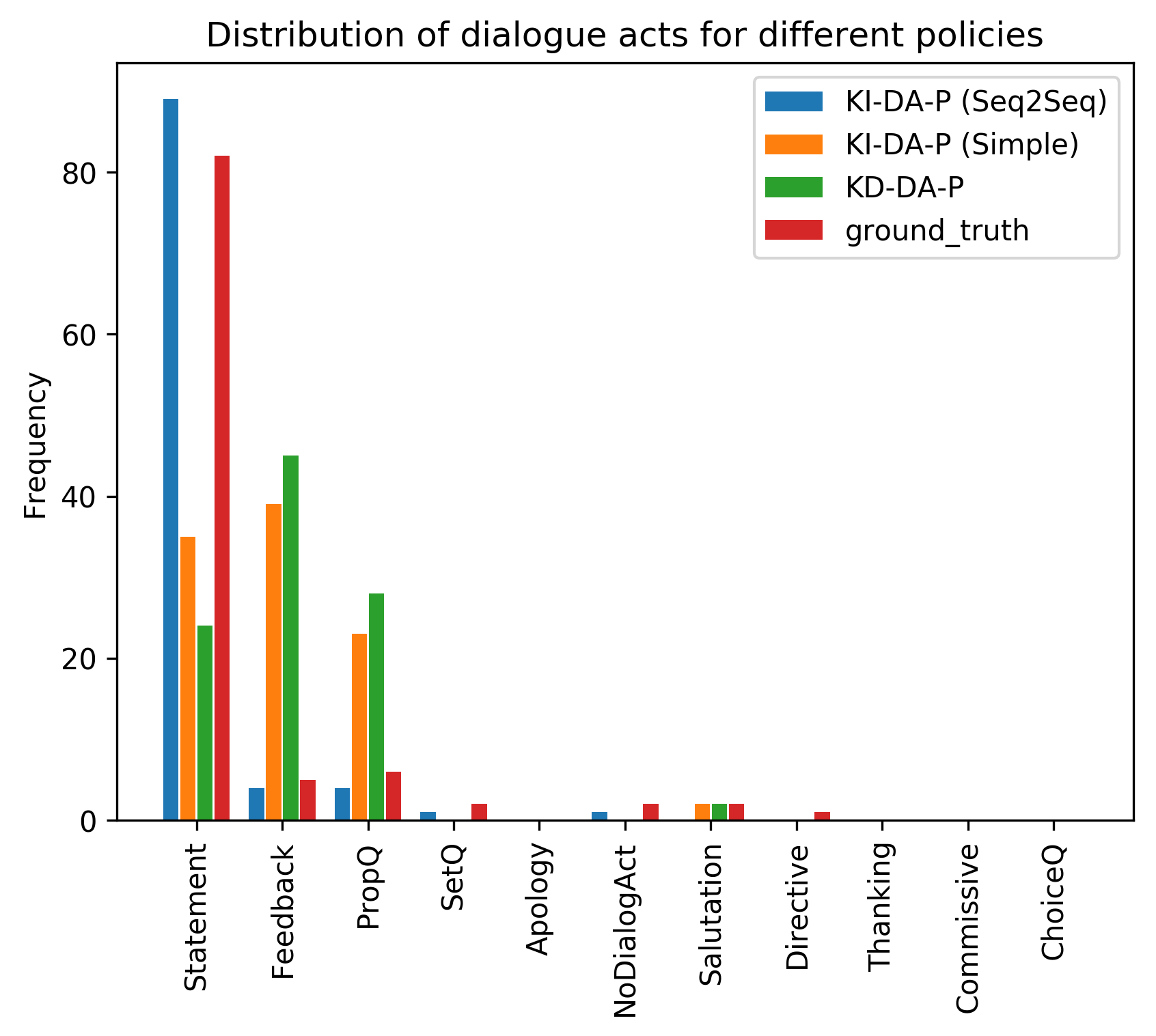}
  		\caption{Distribution of dialogue acts for different dialogue policies}
		\label{fig:differnet_policy}
		  		\vspace*{-2ex}
\end{figure}

Using our dialogue policy models, we estimate an action plan for each turn. Given the dialogue context and the action plan we then generate responses using the PD-NRG w/ DA + knowledge flag model + Past DA model. We evaluate the responses using both automated and human evaluation.
We present our automatic metrics in Table~\ref{auto_predicted_relevant_diversity}. 
The KD-DA-P and KI-DA-P(Simple) produced more~\textit{Feedback} and~\textit{PropQ} dialogue acts than the actual distribution of dialogue acts in the dataset, where most dialogue acts were~\textit{Statements}. We believe this change in the distribution resulted in our models generating responses with fewer words and as a result these models have lower F1-scores. Figure~\ref{fig:differnet_policy} shows the distribution of dialogue acts for different dialogue policies. Multiple knowledge sentences in the unstructured knowledge corpus could be relevant to the dialogue context. When a knowledge sentence that is not the same as the ground truth knowledge is selected, the models could still generate an appropriate response, but n-gram overlap measures will fail to capture their appropriateness. Therefore we limited automated evaluation in this set-up to fewer measures.

For a more realistic comparison of our dialogue policy models to our baselines, we ran human evaluation. We provided a set of crowd workers outputs from two models along with the dialogue context, and asked them ``Which final response is more appropriate for the given conversation?". Crowd workers were provided with 3 options: first response, second response and not sure (limited to those cases when the two responses are equally good/bad). The exact setup given to crowd workers is shown in the Appendix. Table~\ref{human_evaluation_policy_turn} presents results from the manual evaluations. As seen,
the KD-DA-P responses were chosen over the B-Turn model by a large margin. This result is also seen in KD-DA-P responses versus the KI-DA-P (PropQ/AllQ) responses, proving that its is better to have a dialogue policy adapting to the course of the dialogue versus using a fixed distribution~\cite{see2019makes} to predict the dialogue acts. However, the KI-DA-P (Seq2Seq) results in worse responses than the baseline. We believe this is because the~\textit{Statement} dialogue act is a large portion of the dataset, making learning other acts harder for the model. For future work, we will investigate machine learning approaches to learn better models for the dialogue policy. The proposed KD-DA-P results in responses that are better than or similar to human responses in 52\% of the cases. 

\section{Conclusions}
\label{sec:conclusion}

In this work, we propose a policy-driven neural response generation approach for knowledge grounded open-domain dialogue systems. We estimate an action plan that consists of a set of attributes that control the content and style of the generated responses at the turn and sentence levels. We investigate both manual and machine learning based policies. Through human evaluation, we empirically demonstrate that a basic dialogue policy that does sentence level generation outperforms turn level generation, as well as knowledge-grounded response generation baselines. Furthermore, the generated responses realize their respective action plans. This allows builders of dialogue systems control over the model's responses allowing for more consistent user experiences. Our future work includes investigation of better approaches for learning such dialogue policy models along with adding other attributes such as sentiment.


\bibliography{ms}
\bibliographystyle{acl_natbib}

\appendix

\section{Appendices}
\label{sec:appendix}
\subsection{PD-NRG hyperparameters}
\label{sec:hyperparameters}
For training, we fine-tune the GPT double heads model from the HuggingFace repo~\footnote{https://github.com/huggingface/transformers}. The model was initialized with the pre-trained weights. We trained for 3 epochs with a batch size of 2, and 4 distractor responses were used for the next response classification head. We limit the dialogue history to 128 tokens and the knowledge sentence to 32 tokens. Both the language model and classification model tasks had a weight of 1.0. For inference, we use top-\textit{k} and top-\textit{p} sampling with a~\textit{k} value of 0 and~\textit{p} value of 0.9.

\subsection{Policies used}
\begin{figure}[!tbph]
\includegraphics[width=\columnwidth]{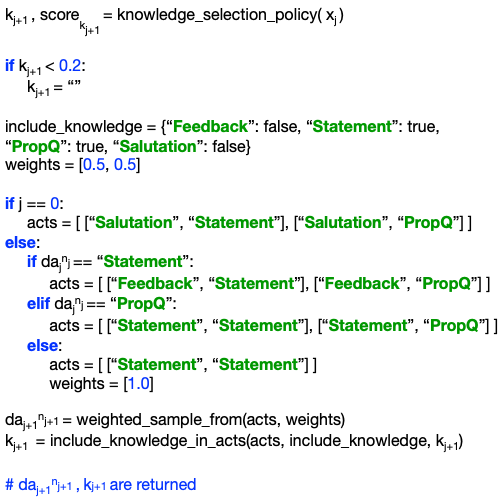}
\caption{Knowledge-Independent DA planning (Simple)}
\label{fig:simple_policy}
\end{figure}

\begin{figure}[h]
\includegraphics[scale=0.4]{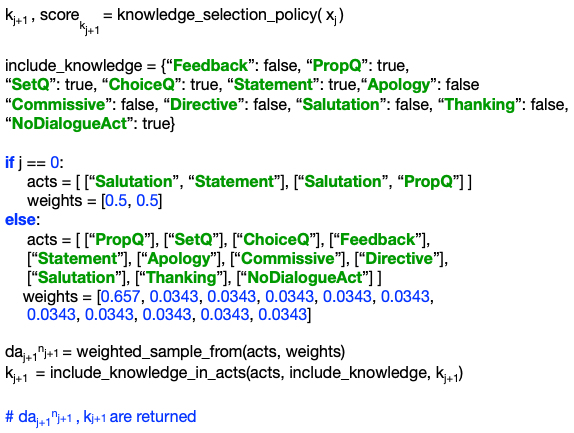}
\caption{Knowledge-Independent DA planning (PropQ)}
\label{fig:propq_policy}
\end{figure}

\begin{figure}[h]
\includegraphics[scale=0.4]{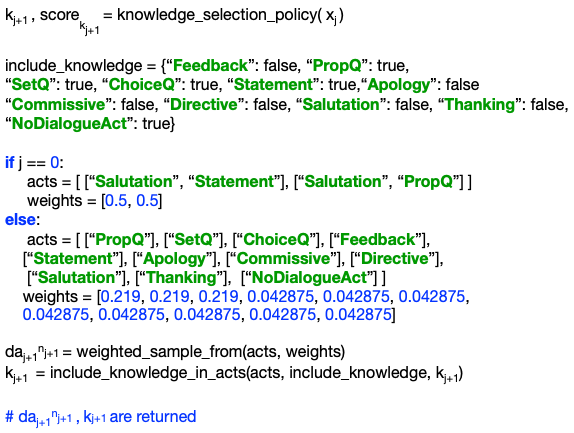}
\caption{Knowledge-Independent DA planning (AllQ)}
\label{fig:allq_policy}
\end{figure}

\begin{figure}[!tbph]
\includegraphics[width=\columnwidth]{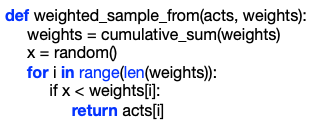}
\caption{Weighted Sample function}
\label{fig:sample_from}
\end{figure}

\begin{figure}[h]
\includegraphics[width=\columnwidth]{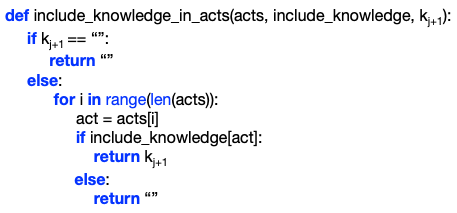}
\caption{Include knowledge function}
\label{fig:include_knowledge}
\end{figure}

\begin{figure}[h]
\includegraphics[width=\columnwidth]{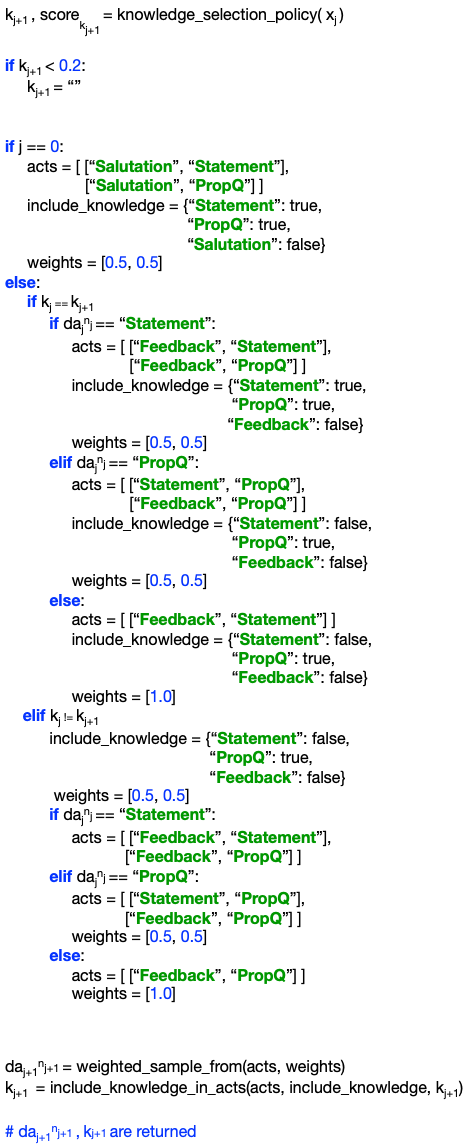}
\caption{Knowledge-Dependent DA planning}
\label{fig:complex_policy}
\end{figure}

Figure~\ref{fig:simple_policy} presents the transitions of the Knowledge-Independent DA planning (Simple). According to this policy, a set of dialogue act sequences are specified based on the last dialogue act of the previous turn. One of these sequences of dialogue acts is chosen (i.e., {\tt weighted\_sample\_from()}) to be included in the action plan. Additionally, the policy determines if the dialogue act should contain the knowledge selected (i.e.,{\tt include\_knowledge\_in\_acts()}). Both methods are defined in Figures~\ref{fig:sample_from} and ~\ref{fig:include_knowledge}. Figure~\ref{fig:complex_policy} presents the transitions for the Knowledge-Dependent DA planning. The transitions are similar to the KI-DA-P (Simple), however in the KD-DA-P one, in addition to the previous dialogue acts, the selected knowledge sentences are also used in the decision when determining the set of dialogue acts for the next turn. Figures~\ref{fig:propq_policy} and~\ref{fig:allq_policy} are two baseline policies that predict question dialogue acts 65.7\% of the time.

\subsection{Layouts for the Crowd Tasks}
\begin{figure}[h]
  		\centering
  		\includegraphics[scale=0.25]{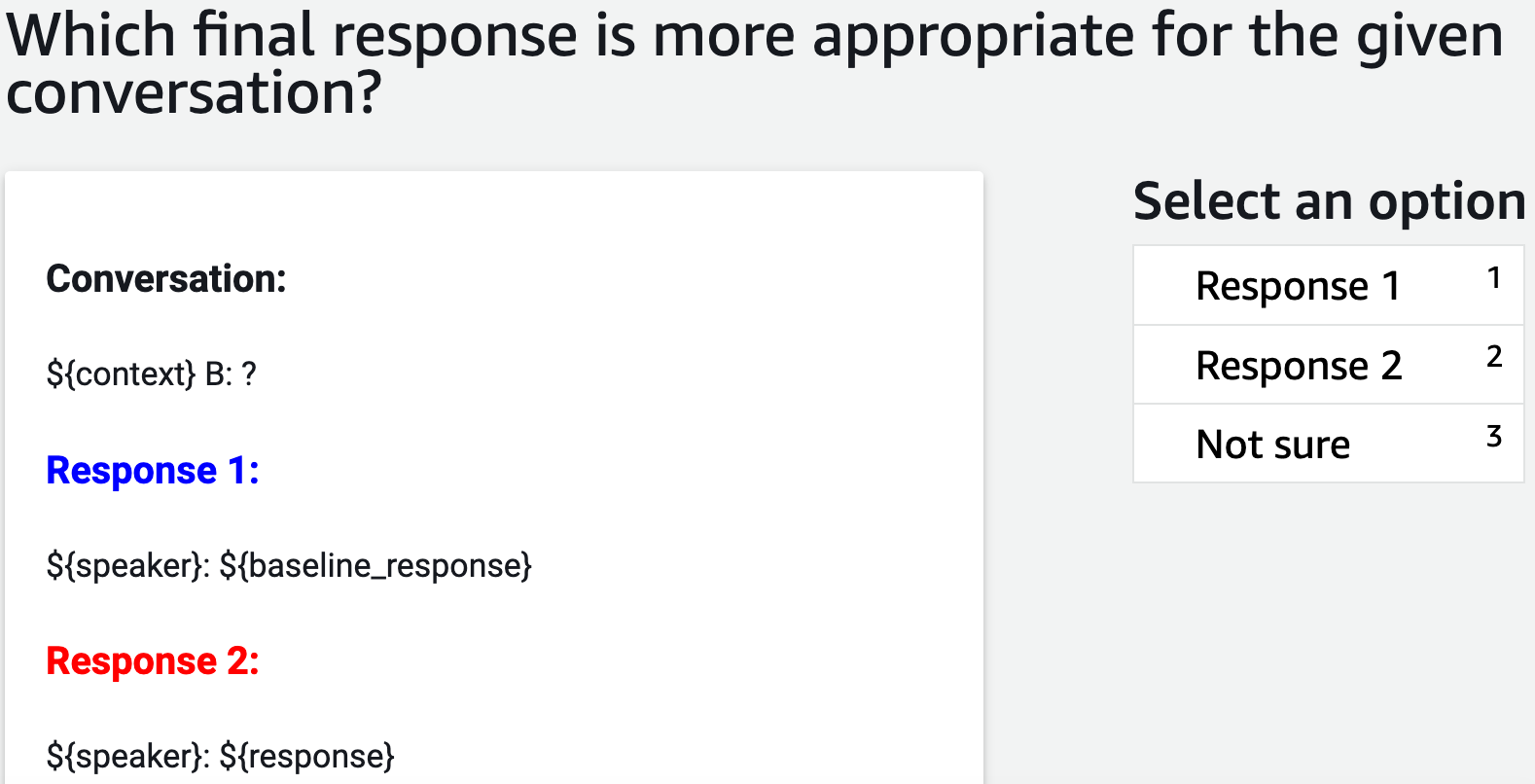}
  		\caption{MTurk layout for Human evaluation}
		\label{fig:human_eval_setup}
\end{figure}

We performed manual evaluations with experienced annotators as well as crowd workers. Figure~\ref{fig:human_eval_setup} shows the interface and instructions we used during manual evalautions.

\end{document}